\documentclass{llncs}
\usepackage{pdf14}

\usepackage{graphicx}

\usepackage{amsmath}
\usepackage{amssymb}
\usepackage{amsfonts}

\usepackage{booktabs}
\usepackage{multirow}
\usepackage{subfig}
\usepackage{cite}
\usepackage{floatflt}
\usepackage{wrapfig}
\usepackage[linesnumbered, ruled,vlined]{algorithm2e}
\usepackage[font=small,skip=0.5em]{caption}

\def\vector#1{\mbox{\boldmath $#1$}}
\newcommand{\CR}{C}

\begin{document}


\title{How \hspace{-0.4em} Far \hspace{-0.4em} Are \hspace{-0.4em} We \hspace{-0.4em} From \hspace{-0.4em} an \hspace{-0.4em} Optimal, \hspace{-0.4em} Adaptive \hspace{-0.4em} DE?}

\author{Ryoji Tanabe\inst{1} \and Alex Fukunaga\inst{2}}
\institute{Institute of Space and Astronautical Science, Japan Aerospace Exploration Agency \and Graduate School of Arts and Sciences, The University of Tokyo}


\maketitle

\begin{abstract}


 We consider how an (almost) optimal parameter adaptation process for an adaptive DE might behave, and compare the behavior and performance of this approximately optimal process to that of existing, adaptive mechanisms for DE.
An optimal parameter adaptation process is an useful notion for analyzing the parameter adaptation methods in adaptive DE as well as other adaptive evolutionary algorithms, but it cannot be known generally.
Thus, we propose a Greedy Approximate Oracle method (GAO) which approximates an optimal parameter adaptation process.
%
We compare the behavior of GAODE, a DE algorithm with GAO, to typical adaptive DEs on six benchmark functions and the BBOB benchmarks, and show that GAO can be used to  (1)  explore how much room for improvement there is in the performance of the adaptive DEs, and (2) obtain hints for developing future, effective parameter adaptation methods for adaptive DEs.

\end{abstract}

\section{Introduction}
\label{sec:introduction}

Differential Evolution (DE) is an Evolutionary Algorithm (EA) that was primarily designed for continuous optimization \cite{StornP97}, and has been applied to many real-world problems \cite{DasMS16}.
A DE population $\vector{P} = \{ \vector{x}^{1}, ..., \vector{x}^{N} \}$ is represented as a set of real parameter vector $\vector{x}^{i} = (x^{i}_{1}, ..., x^{i}_{D})^\mathrm{T}$, $i \in \{1, ..., N\}$, where $D$ is the dimensionality of the target problem and $N$ is the population size.

After initialization of the population, for each generation $t$, for each $\vector{x}^{i,t}$, a mutant vector $\vector{v}^{i,t}$ is generated from the individuals in $\vector{P}^t$ by applying a mutation strategy.
The most commonly used  mutation strategy is the rand/1 strategy:  $\vector{v}^{i,t} = \vector{x}^{r_1,t} + F_{i,t} \: (\vector{x}^{r_2,t} - \vector{x}^{r_3,t})$.
The indices $r_1$, $r_2$, $r_3$ are randomly selected from $\{1, ..., N\}$ such that they differ from each other as well as $i$.
The scale factor $F_{i,t} \in (0,1]$ controls the magnitude of the differential mutation operator.
Then, the mutant vector $\vector{v}^{i,t}$ is crossed with the parent $\vector{x}^{i,t}$ in order to generate a trial vector $\vector{u}^{i,t}$.
Binomial crossover, the most commonly used crossover method in DE, is implemented as follows: For each $j \in \{1, ..., D\}$, if ${\rm rand}[0,1] \leq \CR_{i,t}$ or $j = j_{r}$
(where, ${\rm rand[0,1]}$ denotes a uniformly generated random number from $[0, 1]$, and $j_r$ is a decision variable index which is uniformly randomly selected from $\{1, ..., D\}$),
then $u^{i,t}_j = v^{i,t}_{j}$. Otherwise, $u^{i,t}_j = x^{i,t}_{j}$.
$\CR_{i,t} \in [0,1]$ is the crossover rate.
After all of the trial vectors $\vector{u}^{i,t}$, $i \in \{1, ..., N\}$ have been generated, each individual $\vector{x}^{i,t}$ is compared with its corresponding trial vector $\vector{u}^{i,t}$, keeping the better vector in the population, i.e., if $f(\vector{u}^{i,t}) \leq f(\vector{x}^{i,t})$, $\vector{x}^{i,t+1} = \vector{u}^{i,t}$ for minimization problems. Otherwise, $\vector{x}^{i,t+1} = \vector{x}^{i,t}$.

It is well-known that the performance of EAs is significantly influenced by control parameter settings \cite{EibenHM99,KarafotiasHE15}, and DE is no exception \cite{DasMS16}.
Since identifying optimal control parameter values {\em a priori} is impractical, 
{\em adaptive} DE algorithms, which automatically adjust their control parameters online during the search process, have been studied by many researchers.
Most of the well-known adaptive DEs \cite{BrestGBMZ06, MallipeddiSPT11, ZhangS09, IslamDGRS12, TanabeF13} automatically adjust the $F$ and $\CR$ parameters.
However, while many adaptive DEs have been proposed, their parameter adaptation methods are poorly understood.
Previous work such as  \cite{BrestGBMZ06, MallipeddiSPT11, ZhangS09, IslamDGRS12, TanabeF13} only proposed a novel adaptive DE variant and evaluated its performance on some benchmark functions, but analysis of their  adaptation methods have been minimal.
The situation is not unique to the DE community -- Karafotias et al \cite{KarafotiasHE15} have pointed out the lack of the analysis of adaptation mechanisms in EA. 
There are several previous work that try to analyze the parameter adaptation method in adaptive DE \cite{BrestGBMZ06, ZhangS09, MallipeddiSPT11, DrozdikAAT15, SeguraCSL15}.
However, almost all merely visualized how $F$ and $\CR$ values change during a typical run on  functions, and the analysis is limited to qualitative descriptions such as ``a meta-parameter of $\CR$ in adaptive DE quickly drops down to $[0, 0.2]$ after several generations on the Rastrigin function''. 


In this paper, we consider how an (almost) {\em optimal} parameter adaptation process might behave, and compare the behavior and performance of this approximately optimal process to that of existing, adaptive mechanisms for DE.
We first define what we mean by an optimal parameter adaptation process, and propose a simulation process which can be used in order to greedily approximate the behavior of such an optimal process.
We propose GAODE, which applies this methodology to DE and simulates an approximately optimal parameter adaptation process for a specific adaptive DE framework.
We compare the behavior of GAODE to typical adaptive DE algorithms on six benchmark functions and the BBOB benchmarks \cite{hansen2012fun}, and discuss (1) the performance of current adaptive DE algorithms compared to GAODE, and (2) the implications of these results for developing more effective parameter adaptation method for adaptive DEs.

\section{The proposed GAO framework for adaptive DEs}
\label{sec:approximated_ideal}

First, note that this paper focuses on  {\em parameter adaptation methods} for $F$ and $\CR$ in adaptive DEs such as jDE \cite{BrestGBMZ06}, EPSDE \cite{MallipeddiSPT11}, JADE \cite{ZhangS09}, MDE \cite{IslamDGRS12}, SHADE \cite{TanabeF13}.
In general, the term ``adaptive DE'' denotes a complex algorithm composed of multiple algorithm components.
For example, ``JADE'' consists of three key components: (a) current-to-$p$best/1 mutation strategy, (b) binomial crossover, (c) JADE's parameter adaptation method of $F$ and $\CR$. 
In this paper we want to focus on analyzing (c) {\em only}, rather than ``JADE'', the complex DE algorithm composed of (a), (b) and (c).
Therefore, we extracted only (c) from each adaptive DE variant, and generalized it so that it can be combined with arbitrary mutation and crossover methods.
This approach is taken in recent work \cite{DrozdikAAT15,SeguraCSL15}. 


Due to space limitations, the  parameter adaptation methods in jDE, EPSDE, JADE, MDE, and SHADE cannot be described here (see Section A in the supplemental materials \cite{ppsn16-supplement}), but the general framework can be described as follows:
(i)  At the beginning of each generation $t$, the $F_{i,t}$ and $\CR_{i,t}$ values are assigned to each individual $\vector{x}^{i,t}$. 
 (ii) For each $\vector{x}^{i,t}$, a trial vector $\vector{u}^{i,t}$ is generated using a mutation strategy with $F_{i,t}$ and crossover method with $\CR_{i,t}$. 
(iii) At the end of each generation $t$, the $F$ and $\CR$ values used by successful individuals influence the parameter adaptation on the next generation $t+1$, where we say that an individual $i$
is {\em successful} if $ f(\vector{u}^{i,t}) \leq f(\vector{x}^{i,t})$. 




\subsection{Optimal parameter adaptation process $\vector{\theta}^*$}
\label{sec:ideal_parameter_adaptation}


We define the notion of an optimal parameter adaptation process 
in an adaptive DE. 
Below, DE-$(a,m)$ denotes an adaptive DE algorithm using $a$ and $m$, where 
$a$ is a parameter adaptation method, and $m$ is a DE mutation operator. 
%
Let $L$ be the number of function evaluations (FEvals) until the search finds an optimal solution.
An {\em adaptation process} $\vector{\theta}_m^a = (\{F_1, \CR_1\}, ..., \{F_L, \CR_L\})^{\mathrm T}$ is defined as the series of the $F$ and $\CR$ parameters generated when DE-($a,m$) is executed with some adaptation mechanism $a$ and some DE mutation operator $m$.

For some fixed $m$, an {\em optimal parameter adaptation process} $\vector{\theta}_m^* = (\{F^*_1, \CR^*_1\},$ $...,$ $\{F^*_L, \CR^*_L\})^{\mathrm T}$
is defined as an adaptation process which minimizes the expected value of  $L$, i.e.,  there exists no $a'$ such that $E[|\vector{\theta}_m^{a'}|] < E[|\vector{\theta}_m^*|]$. In the rest of the paper, we abbreviate this as $\vector{\theta}^*$.
An  {\em optimal parameter adaptation method} $a^*$ is an adaptation method such that $\vector{\theta}_m^{a^*} = \vector{\theta}^*$.

$a^*$ and $\vector{\theta}^*$ are useful notions for analyzing the parameter adaptation methods in adaptive DE. 
If $\vector{\theta}^*$  is known for some problem instance $I$,  this by definition is  a lower bound on the performance of DE-$(a, m)$ (no other adaptation process can have a shorter expected length).
This allows quantitative discussions regarding the performance of DE-$(a, m)$ relative to the lower bound, e.g.,  ``DE-(jDE, best/1) is 12.34 times slower than DE-($a^*$, best/1)''.
We can also use such bounds in order to assess whether further improvements to a certain class of methods are worthwhile, e.g.,  ``DE-(JADE, rand/2) performs worse than CMA-ES \cite{HansenK04}, but the performance of DE-($a^*$, rand/2) is better than CMA-ES. Therefore, further improvements to the adaptation method $a$ may result in  a version of DE-($a$, rand/2) which could  outperform CMA-ES.''  


Besides providing a bound  on the performance of DE-$(a, m)$, $\vector{\theta}^*$ might be useful in guiding the development of more efficient parameter adaptation methods.
For example, if for some problem instance $I$,  the $F$ values in $\vector{\theta}^*$ are relatively high at the beginning of the search while they are low at the end of the search, then 
this suggests that  we might be able to  improve the performance of DE-$(a, m)$ on problems similar to $I$ by designing $a$ so that the adaptation process of DE-$(a,m)$ more closely resembles of $\vector{\theta}^*$ for $I$.

However, in practice, it is generally not possible to know $\vector{\theta}^*$.
It is well-known that the appropriate parameter settings depend on the current search situation, and are not fixed values such as $F=0.5$ and $\CR=0.9$, i.e., there are different optimal parameter values as $(\{F^*_1, \CR^*_1\}, \{F^*_2, \CR^*_2\}, \{F^*_3, \CR^*_3\}, ...)$ for each FEvals $(1, 2, 3, ...)$.
$\{ F^*_l, \CR^*_l \}$ are also context-dependent, so we can not compute $\{ F^*_l, \CR^*_l \}$ for some time step $l$ in isolation --  the search state  at $l$ depends on the  control parameter settings used in steps $1,...,l-1$.

\begin{algorithm}[t]
\scriptsize
\SetSideCommentRight
$t \leftarrow 1$, initialize $\vector{P}^t =\{ \vector{x}^{1, t}, ..., \vector{x}^{N, t}\}$, $l \leftarrow 1$, $\vector{\theta}^{\rm GAO} \leftarrow  \emptyset$\;
\While{The termination criteria are not met}{

  \For{$i = 1$ \KwTo $N$}{
    $\vector{U}^l \leftarrow  \emptyset$\; 
    \For{$j = 1$ \KwTo $\lambda$}{
      $F_{l,j} = {\rm rand}(F^{\rm min}, F^{\rm max}]$, $\CR_{l,j} = {\rm rand}[\CR^{\rm min}, \CR^{\rm max}]$\;
    The (virtual) trial vector $\vector{u}^{l,j}$ is generated using an arbitrary mutation strategy with $F_{l,j}$ and crossover method with $\CR_{l,j}$, then $\vector{u}^{l, j} \rightarrow \vector{U}^l$\;
    }
  Evaluates the (virtual) trial vectors in $\vector{U}^l$ by $f$, and select $\vector{u}^{l, {\rm best}}$\; 
    $\vector{u}^{i, t} = \vector{u}^{l, {\rm best}},  \vector{\theta}^{{\rm GAO}} \leftarrow \{F_{l, {\rm best}}, \CR_{l, {\rm best}} \}$, $l \leftarrow l+1$\;
If $f(\vector{u}^{i,t}) \leq f(\vector{x}^{i,t})$, $\vector{x}^{i,t + 1} = \vector{u}^{i,t}$. Otherwise, $\vector{x}^{i,t + 1} = \vector{x}^{i,t}$\;
  }
$t \leftarrow t+1$\;
}
\caption{GAODE (the DE with GAO)}
\label{alg:gaide}
\end{algorithm}

\subsection{Approximating an optimal adaptation process $\vector{\theta}^*$}
\label{sec:gai}

As discussed above, 
$\vector{\theta}^*$ would be very useful for analyzing the parameter adaptation methods, but it cannot be obtained in practice.
Thus, we propose a {\em Greedy Approximate Oracle method} (GAO) in order to approximate $\vector{\theta}^*$, and apply the proposed GAO method to DE. 
%

The basic idea is as follows: suppose that in step (i) of the adaptive DE framework described in the beginning of Section \ref{sec:approximated_ideal}, 
we could enumerate all possible parameter settings $\{F, \CR\}$, and then retroactively select the $\{F, \CR\}$ pair which results in the best child -- this would give us the optimal, 1-step adaptation process.
Similarly, the optimal $k$-step adaptation process can be obtained by recursively simulating the execution of the DE for all possible $k$-step adaptation processes and then selecting the best $k$-step process.
Of course, the number of possible adaptation processes grows exponentially in the number of steps, so in general, the  $k$-step process can not be obtained, and in fact, fully enumerating {\em all} possible 1-step processes is impractical.
We therefore obtain an approximation to the 1-step optimal process by randomly sampling $\{F,\CR \}$ values.


This is implemented as GAODE, shown in Algorithm \ref{alg:gaide}.
For each current FEvals $l$, let us consider that 
the individual $\vector{x}^l$ ($= \vector{x}^{i,t}$) generates the trial vector $\vector{u}^l$  ($= \vector{u}^{i,t}$) using parameter settings $\vector{\theta}^l = \{F_{l}, \CR_{l}\}$ ($=\vector{\theta}^{i,t}$).
The optimal 1-step greedy parameter settings for step $l$ is $\vector{\theta}^{g^*,l} = \{F^{g^*}_{l}, \CR^{g^*}_{l}\}$, and GAO seeks $\vector{\theta}^{{\rm GAO},l}$, which approximates values of $\vector{\theta}^{{g^*},l}$, by random sampling of $\{F,\CR\}$ values.


For each $\vector{x}^l$, $\lambda$ trial vectors $\vector{U}^l =  \{\vector{u}^{l,1}, ..., \vector{u}^{l,\lambda} \}$ are generated  (Algorithm \ref{alg:gaide}, lines $4 \sim 7$).
Parameter values $\vector{\theta}^{l,j} = \{F_{l,j}, \CR_{l,j}\}$, $j \in \{1, ..., \lambda\}$ used for generating $\vector{u}^{l,j}$ are uniformly randomly selected from $(F^{\rm min}, F^{\rm max}]$ and $[\CR^{\rm min}, \CR^{\rm max}]$ respectively  (Algorithm \ref{alg:gaide}, line $6$).
In DE, pseudo-random numbers are used for (a) parent selection in the mutation operator, and (b) the crossover operator.
If two different virtual DE configurations which have different $\{F, \CR \}$ parameter values also use different random numbers for (a) and (b), it complicates the analysis because  we cannot determine whether the configuration which generates the better trial vector did so because of its $\{F, \CR \}$ values or because of the random numbers used in (a) and (b).
Therefore, in our experiments, we synchronized the pseudorandom generators for all of the virtual DE configurations so that they all 
used the same random numbers at both (a) and (b) for generating all trial vectors in $\vector{U}^l$ -- this eliminates the possibility that a virtual DE configuration outperforms another due to fortunate random numbers used for (a) and (b).

The trial vectors in $\vector{U}^l$ are evaluated according to the function $f$, and the  $\vector{u}^{l, {\rm best}}$ with the best (lowest) function value in $\vector{U}^l$ is selected (Algorithm \ref{alg:gaide}, line $8$).
The selected $\vector{u}^{l, {\rm best}}$ is treated as $\vector{u}^l$ ($= \vector{u}^{i,t}$) of $\vector{x}^l$ ($= \vector{x}^{i,t}$).
Note that, $\lambda$ times evaluations according to $f$ which are used to select $\vector{u}^{l, {\rm best}}$ in $\vector{U}^l$ (Algorithm \ref{alg:gaide}, line $8$), are {\em not counted} as the FEvals used in the search -- this simulates a powerful oracle which ``guesses'' $\vector{u}^{l, {\rm best}}$  in one try.
$\vector{\theta}^{l,{\rm best}} = \{F_{l,{\rm best}}, \CR_{l,{\rm best}}\}$ used for generating $\vector{u}^{l, {\rm best}}$ can be considered a approximation to $\vector{\theta}^{g^*,l} = \{F^{g^*}_{l}, \CR^{g^*}_{l}\}$,
and is stored in $\vector{\theta}^{{\rm GAO}}$ (Algorithm \ref{alg:gaide}, line $9$).


Previous work has investigated optimal parameter values in adaptive EAs, especially in Evolution Strategies (ES) community \cite{HansenAA15,EibenHM99,Back93}.
For example, the optimal step size $\sigma^*$ in $(1 + 1)$-ES on the Sphere function is $\sigma^* = 1.224 \: \| \vector{x}^* - \vector{x}  \| / D$ \cite{HansenAA15}, where $\| \vector{x}^* - \vector{x}  \|$ is the Euclidean distance between the optimal solution $\vector{x}^*$ and the current search point $\vector{x}$.
The optimal mutation rate $p_m$ schedule of $(1 + 1)$-GA on the one-max problem is also studied by B{\"{a}}ck \cite{Back93}.
While theoretically well-founded, these results are limited to  a specific algorithm running on a specific problem, and have also been limited to one parameter value, e.g., $\sigma$ and $p_m$.
In contrast, the proposed GAO framework is more general.
While  we focus on applying GAO to DE for black-box optimization benchmarks in this paper,  we believe the  GAO approach can be straightforwardly generalized and applied to combinations of various problem domains (e.g., combinatorial problems, single/multi-objective problems, etc.), algorithms (e.g., GA, ES, MOEA, etc.), and parameters (e.g., crossover and mutation rate, crossover method, etc.).

\section{Evaluating the proposed GAO framework}
\label{sec:experiment}

We compare GAODE, the DE with GAO, to the parameter adaptation methods used by representative adaptive DEs on six benchmark functions.
We show that GAO can be used to  (1)  explore how much room for improvement there is in the  performance of the adaptive DEs (Section \ref{sec:experimental-results-on-sixfunctions}), and (2) obtain hints for developing future, effective parameter adaptation methods (Section  \ref{sec:comp-parameter-adptation-process}).



We used six benchmark functions: Sphere, Ellipsoid, Rotated-Ellipsoid, Rosenbrock, Ackley, Rastrigin functions.
The first three are unimodal, and the last three are multimodal (the Rosenbrock function is unimodal for $D\leq3$).
The Rotated-Ellipsoid and Rosenbrock functions are nonseparable, and the (Rotated-) Ellipsoid functions are ill-conditioned functions.
For details, see Table A.1 in  \cite{ppsn16-supplement}.

The dimensionality $D$ of each function was set to 2, 3, 5, 10, and 20.
The number of runs per problem was 51.
Random number seeds for parts of the DE are synchronized as explained in Section \ref{sec:gai}.
Each run continues until either (i) $|f(\vector{x}^{{\rm bsf}}) - f(\vector{x}^{*})| \leq 10^{-8}$, in which case we treat the run as a ``success'', or (ii) the number of fitness evaluations (FEvals) exceeds $D \times 10^5$, in which case the run is treated as a ``failure''.
$\vector{x}^{{\rm bsf}}$ is the best-so-far solution found during the search process, and $\vector{x}^{*}$ is the optimal solution of the target problem.
Following \cite{HansenK04}, we used the Success Performance 1 (SP1) metric, which is the average FEvals in successful runs divided by the number of successes, as a performance metric of the DE algorithms.
SP1 represents the expected FEvals to reach the optimal solution, i.e., a small SP1 value indicates a fast and stable search.

We used five parameter adaptation methods in the representative adaptive DE variants: jDE \cite{BrestGBMZ06}, EPSDE \cite{MallipeddiSPT11}, JADE \cite{ZhangS09}, MDE \cite{IslamDGRS12}, and SHADE \cite{TanabeF13}.
For details, see Section A in \cite{ppsn16-supplement}.
The most basic rand/1/bin operator \cite{StornP97}, described in Section \ref{sec:introduction}, was used for all DEs.
Following \cite{PosikK12a}, the population size $N$ was set to $5 \times D$ for $D>5$, and $N=20$ for $D = 2$ and $3$.
For each algorithm, we used the control parameter values that were suggested in the original papers as follows: $\tau_{F} = 0.1$ and $ \tau_{\CR} = 0.1$ for jDE, $F$-pool$ = \{0.4, ..., 0.9 \}$ and $\CR$-pool$ = \{0.1, ..., 0.9 \}$ for EPSDE, $c = 0.1$ for JADE, and $H=10$ for SHADE.


In the GAO framework, the parameter generation ranges $(F^{\rm min}, F^{\rm max}]$ and $[\CR^{\rm min}, \CR^{\rm max}]$ have to be set.
In preliminary experiments, GAODE failed on some nonseparable functions when these ranges were set to $(0, 1]$ and $[0, 1]$ respectively.
We believe this failure is due to small $F$ values, so we also evaluated GAODE with  $F^{\rm min} = 0.4$, where $0.4$ is a lowest $F$ value suggested by R\"onkk\"onen et al \cite{RonkkonenKP05}.
Unless explicitly noted, we denote GAODE with the former and later settings as GAODE00 and GAODE04 respectively, and a virtual DE algorithm that is a composition of GAODE00 and GAODE04 as GAODE (GAODE returns the best result obtained by running both GAODE00 and GAODE04).
$\lambda$, the number of configurations sampled by GAODE at each individual,  was set to $200$.

\begin{figure*}[t]
\captionsetup{font=small,skip=0.1em}
\newcommand{\widthvar}{0.9}
\begin{center} 
  \includegraphics[width=\widthvar\linewidth]{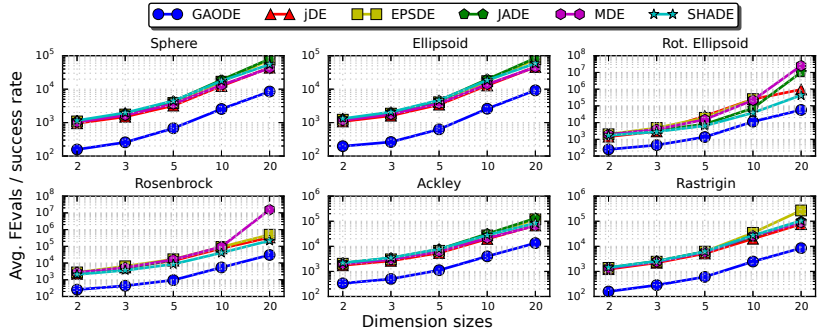}
\caption{
\small
%
Comparison of GAODE with the parameter adaptation methods in the adaptive DEs on each function.
The horizontal axis represents the dimensionality $D$, and the vertical axis represents the SP1 values. 
Data with success rate $= 0$ is not shown.
%
}
\label{fig:sp1_des}
  \end{center}
\end{figure*}

\subsection{Experiment 1: How much room is there for improvement with adaptive DE algorithms with the rand/1/bin operator?}
\label{sec:experimental-results-on-sixfunctions}


Figure \ref{fig:sp1_des} shows the results of GAODE, jDE, EPSDE, JADE, MDE, and SHADE on the six functions.
For GAODE, instead of SP1, we show the lowest FEvals for reaching the optimal solution in the composed results of GAODE00 and GAODE04.
The data of GAODE indicates an approximate bound on the performance that can be obtained by an adaptive DE using the rand/1/bin operator.



As shown in Figure \ref{fig:sp1_des}, all runs of EPSDE fail on the Rotated-Ellipsoid function for $D=20$.
JADE also fails on all runs on the Rosenbrock function when  $D \geq 10$.
MDE can reach the optimal solution on the both functions, but its SP1 values are significantly worse than other methods.
Consistent with the results in  \cite{ZhangS09,SeguraCSL15}, 
adaptation methods tend to perform poorly when used with operators that are different from the operators used in the original papers where the adaptation methods were proposed.
Although the performance rank among the methods depends on the functions and the dimensionality $D$, 
jDE and SHADE perform better than other compared methods in almost all cases.
However, as shown in Figure \ref{fig:sp1_des}, jDE and SHADE converge to the optimal solution $4 \sim 20$ times slower than GAODE. 
This shows that even the best current adaptive methods perform poorly compared to an approximation of a 1-step greedy optimal process (and are therefore even worse compared to a $k$-step optimal process).
{\em Thus, it appears that despite significant progress in recent years, there is still significant room for improvement in parameter adaptation methods for DE}.


\subsection{Experiment 2: How should we adapt the control parameters?}
\label{sec:comp-parameter-adptation-process}

\begin{figure*}[t]
\small
\captionsetup[subfloat]{font=scriptsize,captionskip=0.1em}
\captionsetup{font=small,skip=0.1em}
\newcommand{\widthvar}{0.24}
\begin{center} 
  \subfloat[Rosenbrock ($D=10$)]{
    \label{fig:heatmap_shade_D10}
    \includegraphics[width=\widthvar\linewidth]{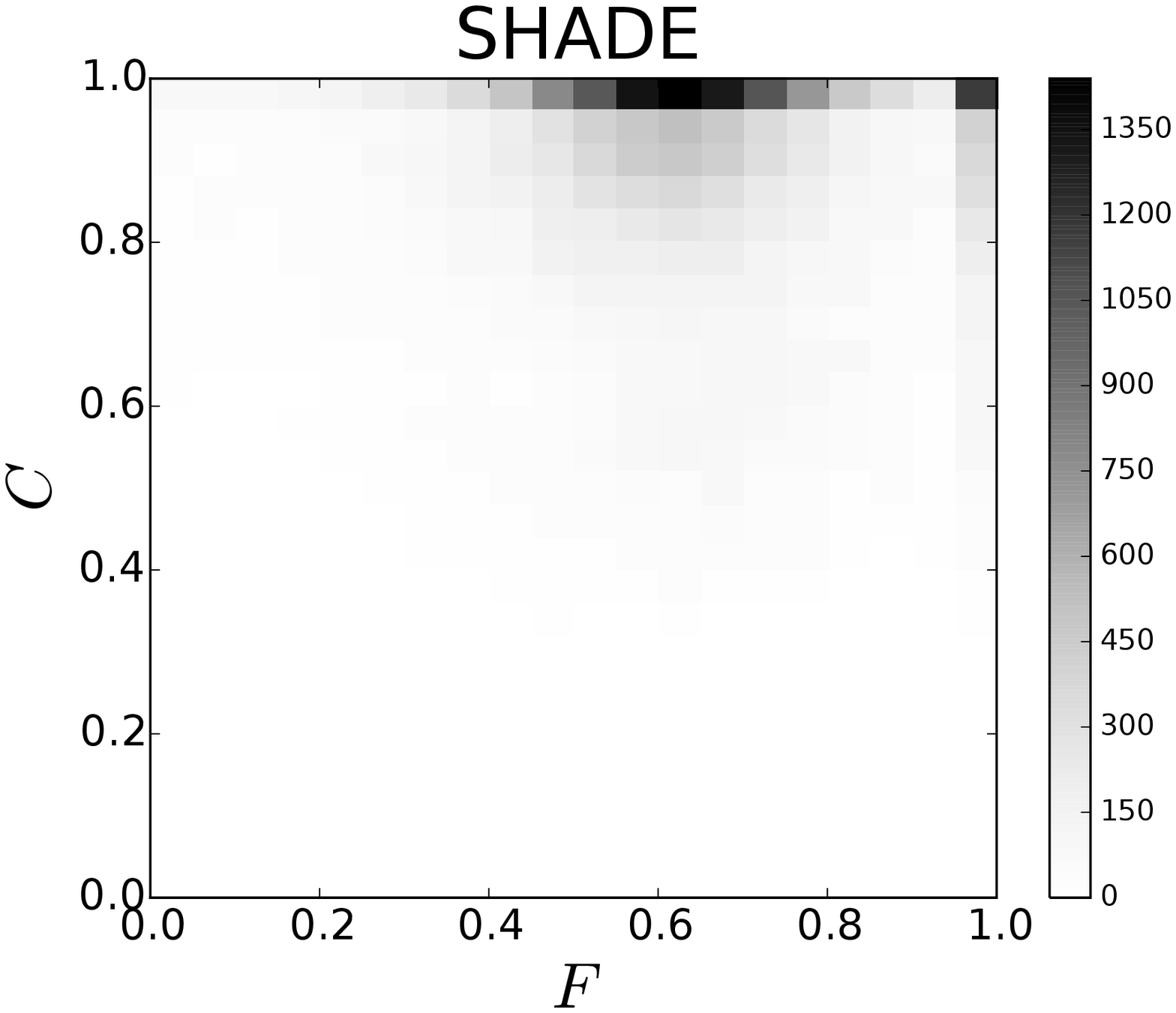}
    \includegraphics[width=\widthvar\linewidth]{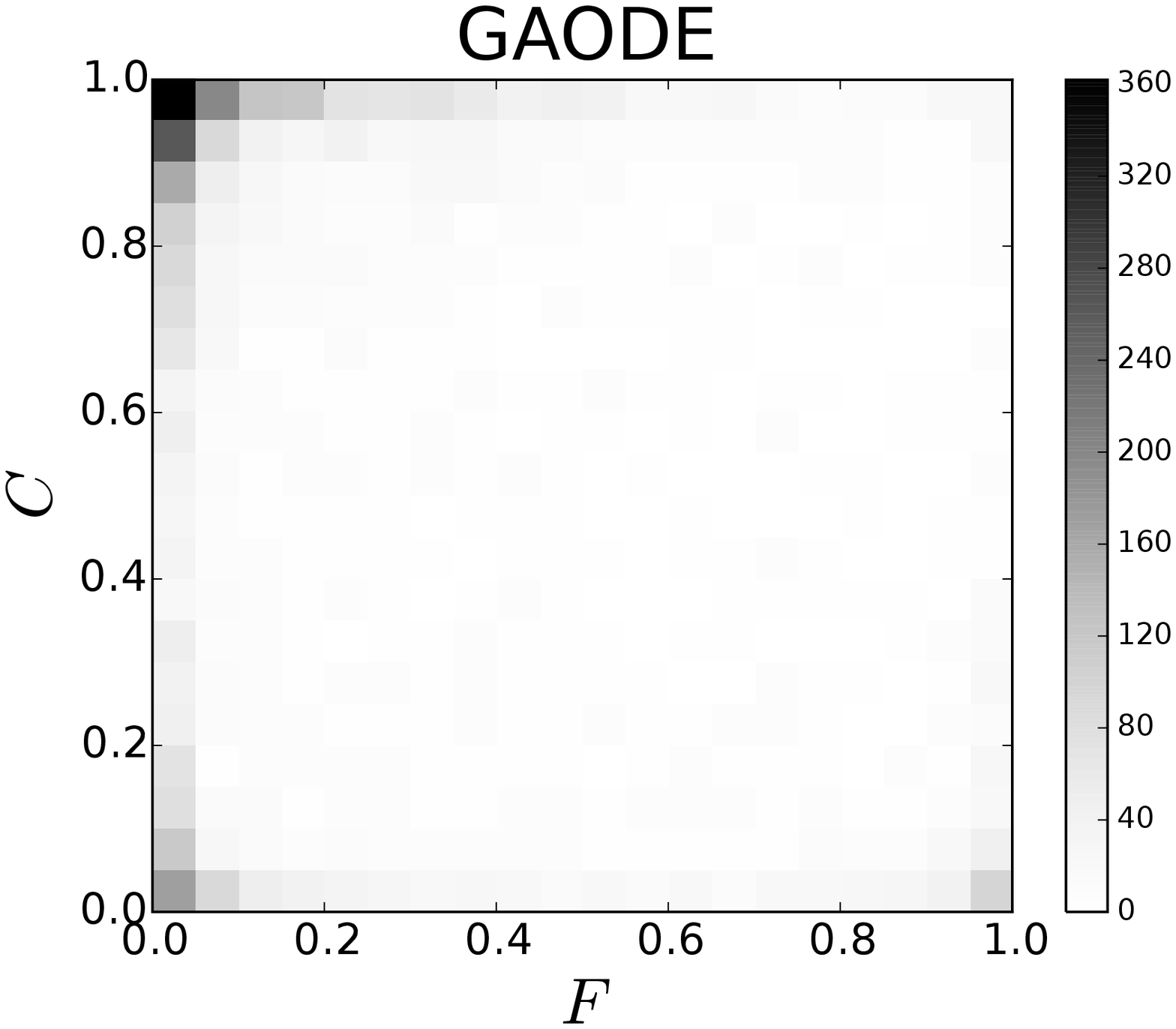}
  }
  \subfloat[Rastrigin ($D=10$)]{
    \label{fig:heatmap_idealde_D10}
    \includegraphics[width=\widthvar\linewidth]{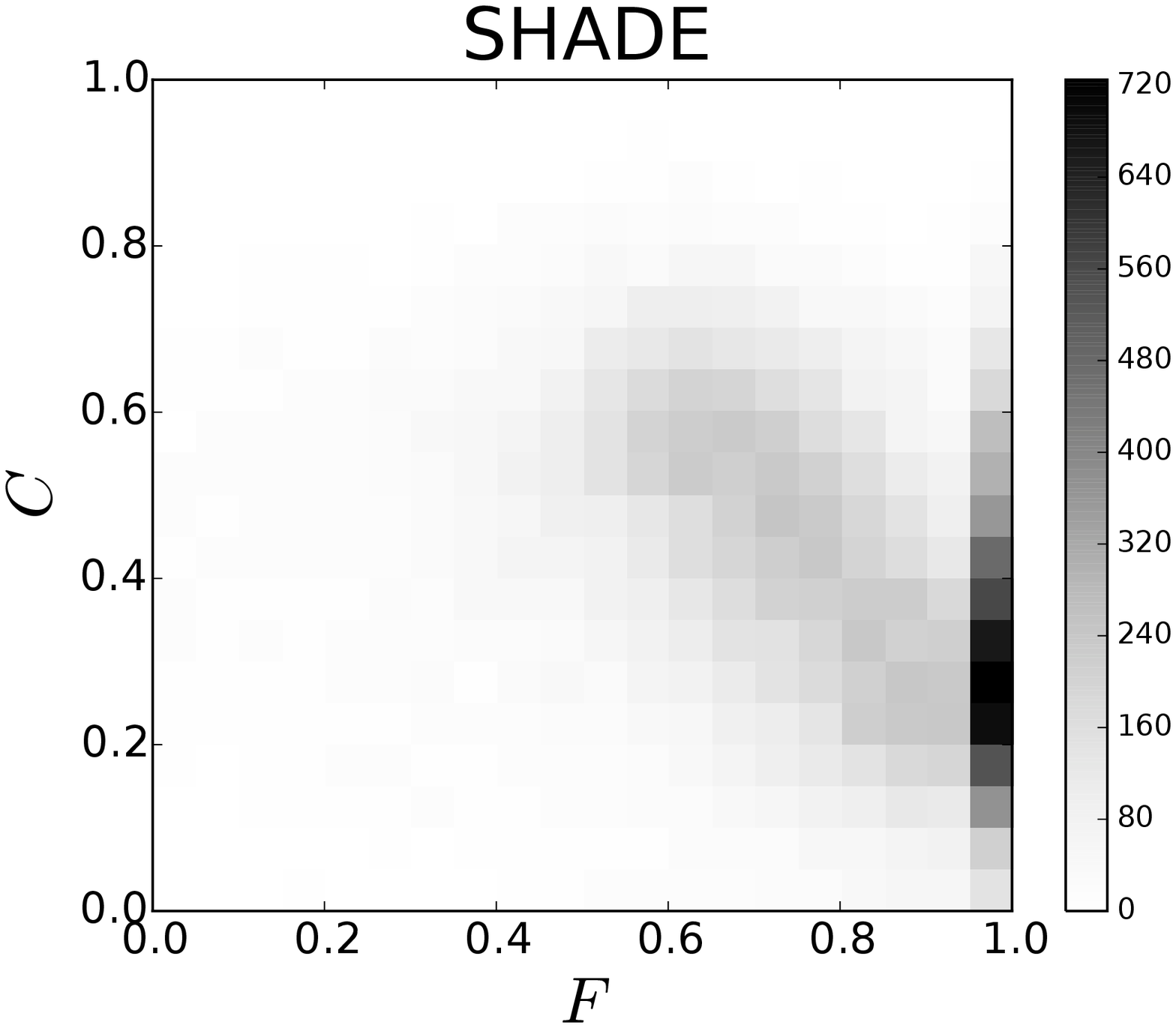}
    \includegraphics[width=\widthvar\linewidth]{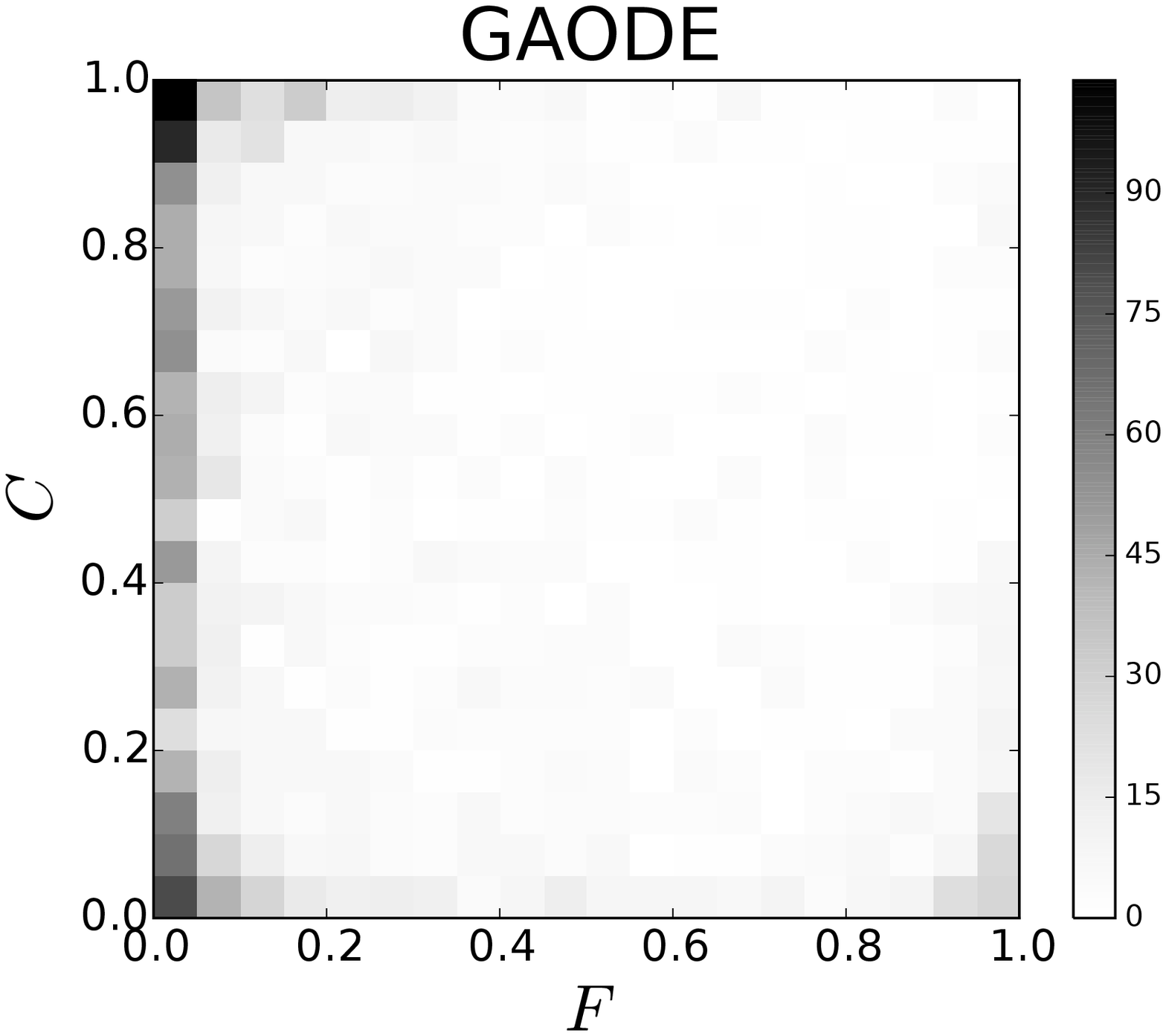}
  }
\caption{
\small
Frequency of appearance of  $\{F, \CR\}$ value pairs during the search process for SHADE and GAODE on the 10-dimensional (a) Rosenbrock and (b) Rastrigin functions.
Darker colors indicate more frequent generation of the corresponding values by the parameter adaptation method.
%
}
\label{fig:heatmap}
  \end{center}
\end{figure*}

Let us consider how the behavior of GAODE differs from existing adaptation methods.
Figure \ref{fig:heatmap} shows the frequency of appearance of  $\{F, \CR\}$ value pairs during the search process for SHADE and GAODE on the 10-dimensional Rosenbrock and Rastrigin functions.
Data from the best run out of 51 runs is shown.
The results of jDE, EPSDE, JADE, and MDE can be seen in Figure A.1 in \cite{ppsn16-supplement}. 


As shown in Figure \ref{fig:heatmap}, SHADE frequently generates $F$ and $\CR$ values in the range $[0.5, 0.7]$ and $[0.9, 1.0]$ on the Rosenbrock function, and  $[0.9, 1.0]$ and $[0.1, 0.4]$ on the Rastrigin function respectively.
These results are consistent with previous studies for DE \cite{BrestGBMZ06, DasMS16} and adaptive DEs \cite{BrestGBMZ06, ZhangS09}.
On the other hand, GAODE mainly generates $F$ values in the range $[0.0, 0.1]$ on both functions.
The $\CR$ values frequently appear in $[0.0, 0.2]$ and $[0.8, 1.0]$ on the Rastrigin function, and GAODE mainly generates $\CR$ values in {\em both} $[0.9, 1.0]$, and $[0.0, 0.1]$ on the Rosenbrock function, i.e., the $\CR$ values are bimodal.
Interestingly, for the both functions, GAODE occasionally generates $F$ and $\CR$ values in the extreme regions $[0.9, 1.0]$ and $[0.0, 0.1]$ respectively (see bottom right in the figures).

In summary, GAODE frequently generates  small $F$ values, and $\CR$ values in the range $[0, 0.2]$ and $[0.8, 1]$.
Although CoBiDE \cite{WangLHL14}, a recently proposed {\em non-adaptive} DE,  generates the $F_{i,t}$ and $\CR_{i,t}$ values for each $\vector{x}^{i,t}$ according to a bimodal (two Cauchy) distribution, we are not aware of such a bimodal  sampling approach in any previously proposed adaptive method. 
An adaptive DE algorithm using such sampling method may also perform better than the existing methods \cite{BrestGBMZ06, MallipeddiSPT11, ZhangS09, IslamDGRS12, TanabeF13}.
Thus, analysis of the approximate optimal parameter adaptation process obtained by GAO suggests that instead of unimodal sampling procedures implemented in previous adaptation methods, adaptive mechanisms using multimodal sampling  may be a promising direction for future work.



\section{Comparing GAODE with state-of-the-art EAs}
\label{sec:bbob_results}

GAODE, which is an approximate simulation of an optimal, 1-step adaptation process, 
significantly performs better than the current state-of-the-art parameter adaptation methods for DE using the rand/1/bin operator, as described in Section \ref{sec:experimental-results-on-sixfunctions} (again, we reemphasize that {\em GAODE is not a practical algorithm and is for analysis only} -- the ``performance'' of GAODE ignores the $\lambda - 1$ samples which are discarded by GAODE at each iteration).
It is interesting to compare GAODE with other state-of-the-art EAs.
Here, we compare the adaptive DE variants including GAODE\footnote{
The BBOB benchmarks provide 15 instances for each function, i.e., there are $24 \times 15 = 360$ function instances.
In this study, we applied GAODE00 and GAODE04 three times for each instance, and only the best result among them is used for GAODE.}
 with HCMA \cite{LoshchilovSS13a} and best-2009  on the BBOB benchmarks, consisting of 24 various functions \cite{hansen2012fun}.
HCMA, an efficient surrogate-assisted algorithm portfolio, represents the state-of-the-art on the BBOB benchmarks.
Best-2009 is a {\em virtual} algorithm portfolio that is retrospectively constructed from the performance data of 31 algorithms participating in the GECCO BBOB 2009 workshop.
Is it possible for an adaptive DE algorithm using the {\em classical} rand/1/bin operator to be competitive with these methods?

\begin{figure*}[t]
\small
\newcommand{\widthvar}{0.32}
  \begin{center} 
\includegraphics[width=\widthvar\textwidth]{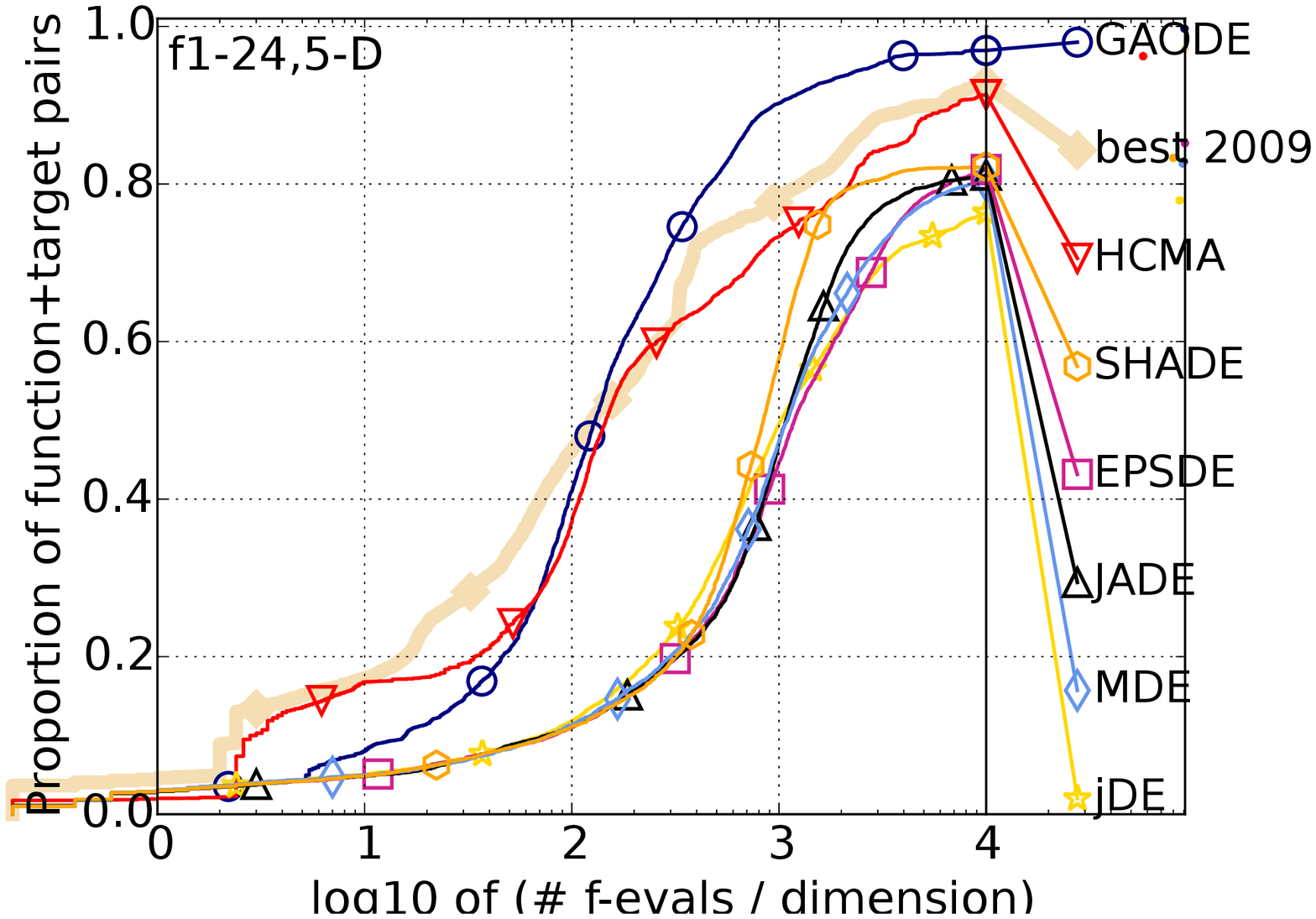}
\includegraphics[width=\widthvar\textwidth]{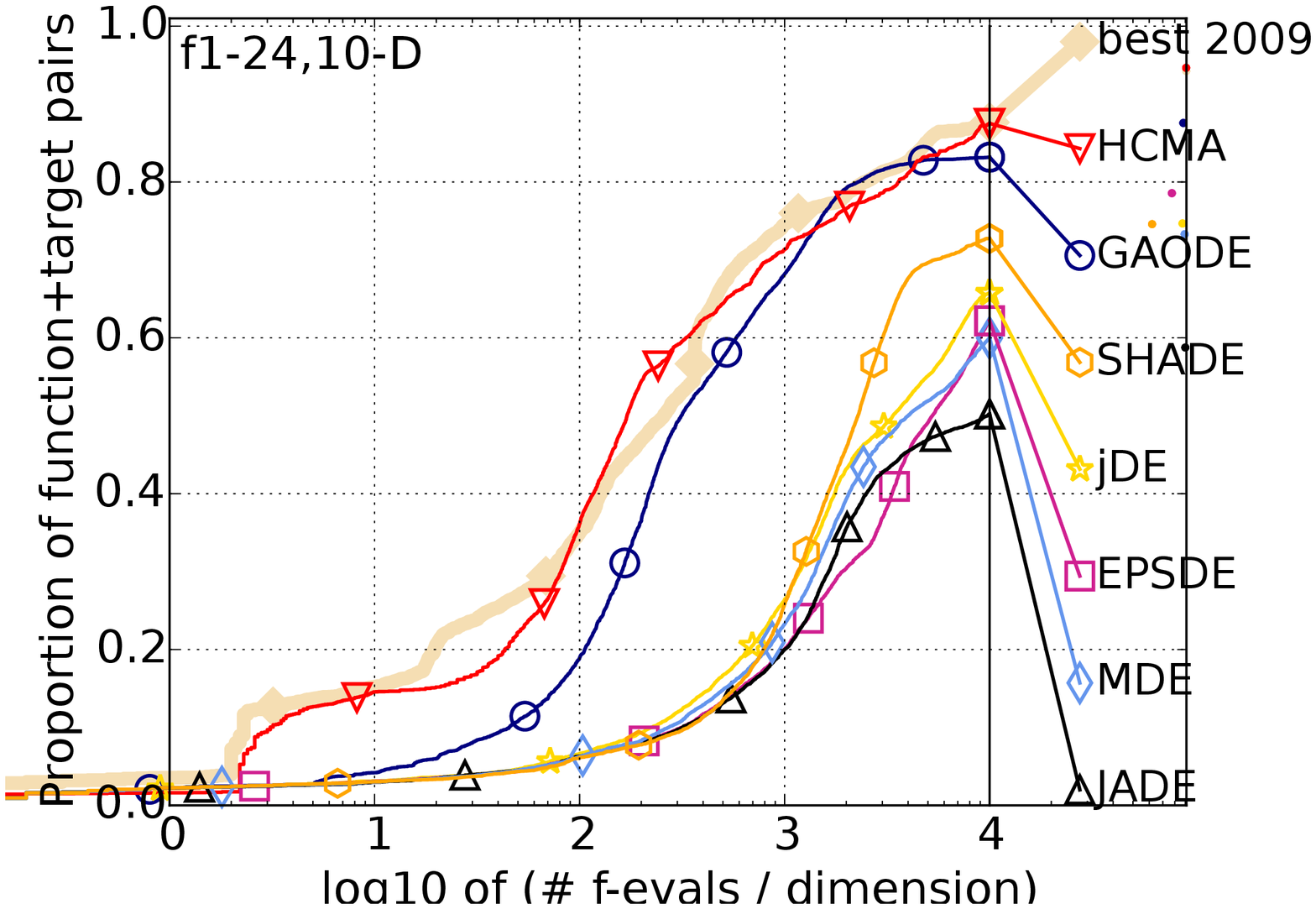}
\includegraphics[width=\widthvar\textwidth]{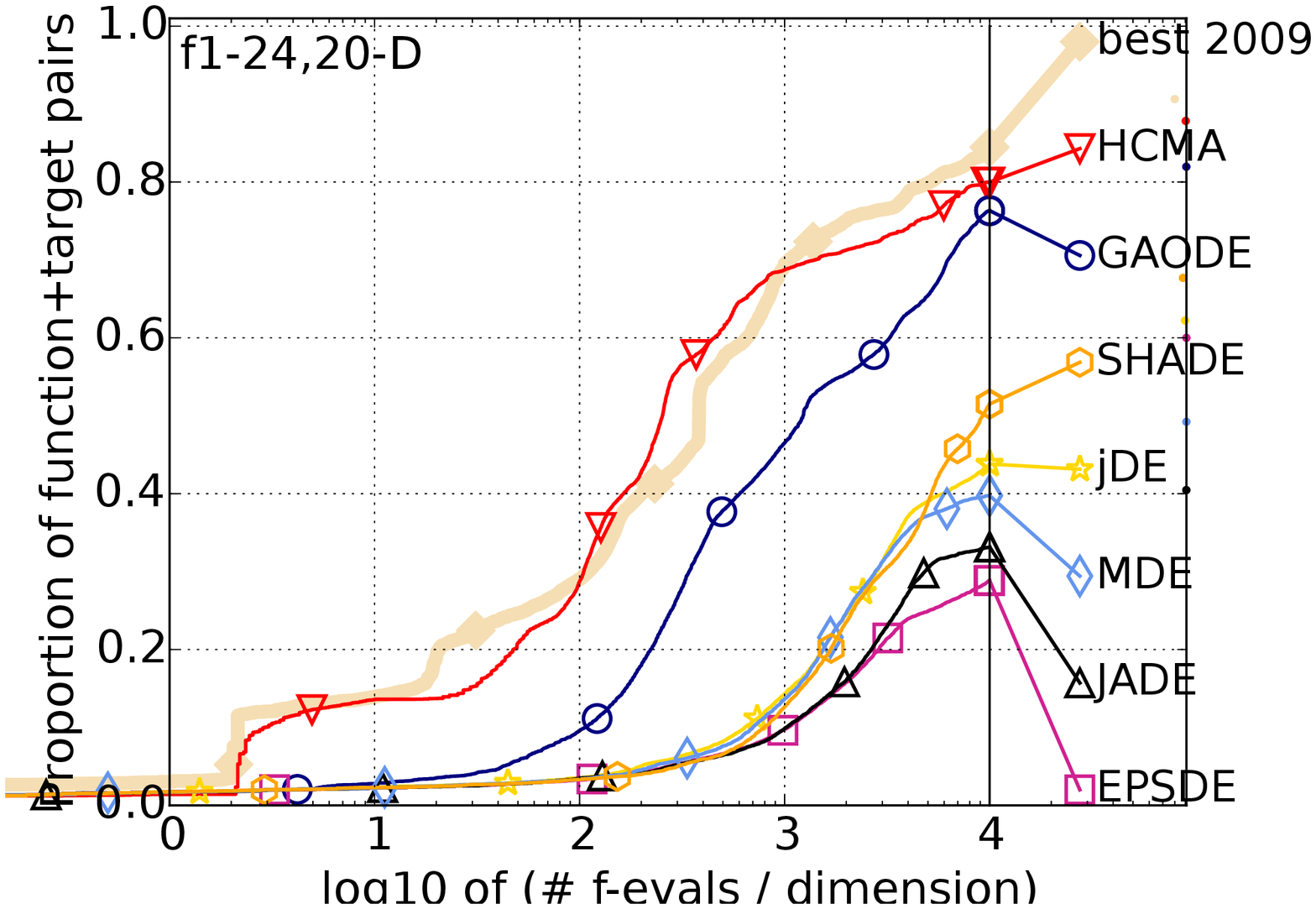}
\caption{
\small
Comparisons of GAODE with the adaptive DE variants, HCMA, and best-2009 on BBOB benchmarks ($D \in \{5, 10, 20 \}$).
These figures show the bootstrapped Empirical Cumulative Distribution Function (ECDF) of the FEvals divided by dimension for 50 targets in $10^{[-8..2]}$ for 5, 10, 20 dimensional all functions (higher is better).
For details of the ECDF, see a manual of \texttt{COCO software} (\url{http://coco.gforge.inria.fr/}).
}
\label{fig:bbob_results}
  \end{center}
\end{figure*}

Figure \ref{fig:bbob_results} shows the Empirical Cumulative Distribution Function (ECDF) for each algorithm for 24 BBOB benchmark problems ($D =5, 10, 20$) when maximum FEvals $= D \times 10^4$.
The results for each function class and for $D= 2, 3$ can be found in Figures A.2 -- A.6 in  \cite{ppsn16-supplement}.
As shown in Figure \ref{fig:bbob_results}, GAODE clearly outperforms jDE, EPSDE, JADE, MDE, and SHADE for all dimensions, in terms of both the quality of the best-so-far solution obtained during the search process and the anytime performance.
GAODE also performs significantly better than HCMA and best-2009 for $D \leq 5$.
This result suggests that if we can find a parameter adaptation method which performs similarly to the GAODE model, then an adaptive DE algorithm using the classical rand/1/bin could possibly outperform state-of-the-art algorithm portfolios such as HCMA for $D \leq 5$.

On the other hand, when the dimensionality increases, the performance of GAODE degrades compared to HCMA and best-2009.
For $D=20$, GAODE is outperformed by HCMA and best-2009.
This may indicate that for high-dimensionality problems, it may not be possible to develop an adaptive DE using the rand/1/bin operator which is competitive with methods such as HCMA.
However, this result may be due to the fact that GAODE only simulates an approximately optimal 1-step adaptation process -- increasing the number of steps (i.e., a $k$-step optimal process) may result in better results, and is a direction for future work. In addition, different mutation operators (e.g., best/2, current-to-$p$best/1, etc.) may enable significantly better performance for adaptive DEs.

\section{Conclusion}
\label{sec:conclusion}


We proposed a Greedy Approximate Oracle method (GAO) which approximates an optimal parameter adaptation process $\vector{\theta}^*$. 
In GAO, $\lambda$  parameter sets are randomly generated for each individual in the population, and the best parameter set  with respect to the objective function value is used as a greedily approximated optimal parameter set (the other $\lambda -1$ sets are discarded and are not counted).
We evaluated GAODE, a DE algorithm with GAO, on 6 standard benchmark functions and the BBOB benchmarks \cite{hansen2012fun}, and compared it with the parameter adaptation methods in 5 adaptive DE variants.
We showed that  (1)  current adaptive DEs are significantly worse than even an approximate,  1-step optimal adaptation, suggesting that there is still much work to be done in the development of adaptive mechanisms (Section \ref{sec:experimental-results-on-sixfunctions}), and (2) GAO can be used to identify promising  directions for developing an efficient parameter adaptation method in adaptive DE (Section \ref{sec:comp-parameter-adptation-process}).
We also compared GAODE with HCMA \cite{LoshchilovSS13a} and best-2009 on the BBOB benchmarks  \cite{hansen2012fun} in Section \ref{sec:bbob_results}, and showed that a better adaptive mechanism may enable a DE using the classical rand/1/bin operator to achieve state-of-the-art performance. 



The proposed GAO framework is a first attempt to approximate the optimal parameter adaptation process, and there is much room for improvement, as discussed in Section \ref{sec:approximated_ideal}. 
In this paper, we applied GAO to the DE with the rand/1/bin operator, and evaluated its performance on single-objective continuous optimization problems.
Future work will explore GAO as a general framework that can be applied to analyze the behavior of any adaptive EA (independent of specific operators and problem domains).


\bibliographystyle{abbrv}
\bibliography{reference}

\clearpage




\end{document}